\documentclass[sn-chicago,pdflatex]{sn-jnl}
\usepackage{svg}
\usepackage{lmodern}
\usepackage{anyfontsize}
\usepackage{graphicx}
\usepackage{multirow}
\usepackage{amsmath,amssymb,amsfonts}
\usepackage{amsthm}
\usepackage{mathrsfs}
\usepackage[title]{appendix}
\usepackage{xcolor}
\usepackage{textcomp}
\usepackage{manyfoot}
\usepackage{booktabs}
\usepackage[linesnumbered,ruled,noend]{algorithm2e}
\usepackage{listings}
\usepackage[caption=false]{subfig}

\newtheorem{definition}{Definition}

\newcommand{\castor}[0]{\textsc{Castor}}

\DeclareMathOperator*{\argmax}{arg\,max}
\DeclareMathOperator*{\argmin}{arg\,min}

\SetKwInOut{Input}{Input}
\SetKwInOut{Output}{Output}
\SetEndCharOfAlgoLine{}
\SetKwComment{Comment}{\# }
\SetCommentSty{\footnotesize}

\begin{document}

\title[\castor{}]{\castor{}: Competing shapelets for fast and accurate time series classification}

\author*[1]{\fnm{Isak} \sur{Samsten}}\email{samsten@dsv.su.se}

\author[1]{\fnm{Zed} \sur{Lee}}\email{zed.lee@dsv.su.se}


\affil*[1]{\orgdiv{Department of Computer and System Sciences},
	\orgname{Stockholm University}, \orgaddress{\street{Borgarfjordsgatan 12}, \city{Kista},
		\postcode{16455}, \state{Stockholm}, \country{Sweden}}}

\abstract{ %
	Shapelets are discriminative subsequences, originally embedded in shapelet-based decision trees but have since been extended to shapelet-based transformations.
	We propose \castor{}, a simple, efficient, and accurate time series classification algorithm that utilizes shapelets to transform time series.
	The transformation organizes shapelets into groups with varying dilation and allows the shapelets to \emph{compete} over the time context to construct a diverse feature representation.
	By organizing the shapelets into groups, we enable the transformation to transition between levels of competition, resulting in methods that more closely resemble distance-based transformations or dictionary-based transformations.
	We demonstrate, through an extensive empirical investigation, that \castor{} yields transformations that result in classifiers that are significantly more accurate than several state-of-the-art classifiers.
	In an extensive ablation study, we examine the effect of choosing hyperparameters and suggest accurate and efficient default values.
}


\maketitle

\section{Introduction}\label{sec:introduction}

A time series is a sequentially ordered collection of values. Time series analysis includes broad tasks such as similarity search \citep{sim2022revie}, anomaly detection \citep{blazquez-garcia2022revie}, forecasting and nowcasting \citep{lim2021times}, clustering \citep{holder2024revie}, classification \citep{bagnall2017great}, and generative modeling \citep{brophy2023gener}. In this paper, we explore time series classification, where the objective is to construct a model that can, given a collection of labeled time series, assign the correct label to a previously unseen time series. Time series classification has applications in numerous domains, such as identifying abnormal electrocardiograms \citep{chauhan2015anomaly} or classifying insects based on their sound profiles \citep{petitjean2016faster}.

The main goal of time series classification is to obtain good predictive performance, usually measured by how accurately a model predicts the correct label (i.e., \textit{classification accuracy}). Currently, the most accurate time series classification methods are based on various model features such as random convolution in Rocket, MultiRocket, and Hydra \citep{dempster2020rocke,dempster2023hydra}, discriminative subsequences (i.e., \textit{shapelets}) as in dilated shapelet transform (DST) \citep{guillaume2022randp}, phase-independent dictionaries as in bag-of-SFA-symbols (BOSS) and Weasel \citep{schafer2015boss,schafer2023wease}, or diverse ensembles as in HIVE-COTE \citep{middlehurst2021hive-}, where each method has different trade-off ratio between classification accuracy and computational cost.

The concept of shapelets, which are discriminative subsequences of time series, was first introduced by \citet{ye2009time} and has undergone several enhancements to improve classification accuracy \citep{lines2012shape,grabocka2014learn,guillaume2022randp}, to reduce computational cost \citep{rakthanmanon2013fast}, or to achieve both \citep{wistuba2015ultra,karlsson2016gener}. In addition to continuous improvements in accuracy and computational efficiency, shapelet-based methods offer another significant advantage over the other methods: the resulting models use inherently explainable components extracted from the training data. Whether through model-specific or post-hoc methods, shapelets provide the basis to explain the model's performance or the rationale behind a specific label assignment. Another advantage of shapelets over random convolution kernels is that they constitute actual time series characteristics, allowing for more accurate classification models to be constructed. For instance, despite constructing similar feature representations, DST is significantly more accurate than Rocket \citep{guillaume2022randp} with the same number of shapelets and kernels.

One common characteristic of the most successful time series classifiers is the use of large, size-controlled feature spaces that capture several thousand features for each time series and then construct simple linear models \citep[][see, e.g.,]{tan2022multi}. Typically, these large feature spaces are constructed using randomized patterns and hyperparameters, resulting in transformations that are fast to compute and yield models with high predictive performance \citep{guillaume2022randp,dempster2020rocke}. A significant innovation is the use of \emph{dilation}, whereby gaps are introduced into the patterns, such that \emph{convolutions} \citep{dempster2020rocke} or \emph{distance profiles} \citep{guillaume2022randp} incorporate an extended temporal context for feature generation.

In this work, we introduce \castor{}, a novel supervised shapelet-based time series transformation that can be used for constructing extremely accurate classification models. \castor{} incorporates both directions of dilation (i.e., convolutions and distance profiles) and provides a hybrid of shapelet- and dictionary-based methods. Like DST, \castor{} transforms time series the distance profile of shapelets but arranges the shapelets into groups and forces the shapelets in each group to compete over the temporal contexts. In our experiments, we show that \castor{} significantly outperforms both the current state-of-the-art classifiers using random convolution, such as Rocket \citep{dempster2020rocke} and Hydra \citep{dempster2023hydra}, as well as all shapelet-based competitors in terms of predictive performance, resulting one of the most accurate classifiers for time series. In Section~\ref{sec:ablation}, we show that allowing the shapelets to \emph{compete} over temporal contexts is the primary driver of the superior accuracy of \castor{}.

\smallskip\noindent The main contributions of \castor{} are as follows:

\begin{itemize} 
    \item \textbf{Effectiveness.} \castor{} achieves the highest accuracy among shapelet-based time series classifiers, offering state-of-the-art predictive performance with minimal parameter tuning and feature selection. It outperforms most state-of-the-art classifiers, including Rocket and MultiRocket. We also demonstrate a trade-off between competition and independence that influences predictive performance.
    \item \textbf{Efficiency}: Utilizing the same number of features as comparable classifiers such as Hydra, \castor{} demonstrates superior runtime efficiency compared to state-of-the-art random convolution-based classifiers like Rocket and MultiRocket while maintaining comparable predictive performance. Notably, \castor{} can train and test on 112 datasets from the UCR time series repository approximately 20\% faster than Rocket, 30\% faster than MultiRocket, and 60\% faster than DST. 
    \item \textbf{Completeness.} We perform a comprehensive ablation study to examine the different algorithmic options, emphasizing the significance of \emph{competition} among shapelets and \emph{diversity} in the transformation process. While \castor{} possesses multiple hyperparameters, we propose default settings for these parameters and demonstrate that they yield predictive performance superior to most state-of-the-art classifiers.
    \item \textbf{Reproducibility.} We provide the community with an efficient implementation featuring a standardized Python interface, ensuring ease of reproduction and usage.
\end{itemize}

The remainder of this paper is structured as follows: Section~\ref{sec:related-work} provides a review of the related literature. Section~\ref{sec:methods} details the \castor{} framework and discusses the algorithmic decisions made. Section~\ref{sec:results} reports on the experimental evaluation of both predictive accuracy and computational efficiency and includes a comprehensive ablation study of the hyperparameters.

\section{Related work}\label{sec:related-work}
\subsection{Shapelet-based classification methods}

The initial idea involves considering every subsequence of every time series in an input dataset and assessing their discriminatory power using a scoring function.
\citet{ye2009time} employs the information gain measure and incorporates subsequence extraction into a decision tree algorithm \citep{gordon1984class}.
Other measures have been investigated to enhance predictive performance, such as the F-statistic, Kruskal-Wallis, and Mood's median \citep{lines2012shape,hills2014class}.
Rather than embedding subsequence extraction in a decision tree, \citet{lines2012shape} investigates a shapelet-based transform (ST), which enables the use of any classifier for the subsequent classification task. In this approach, the most discriminative shapelets are extracted, and the resulting transformation comprises the minimum distance from a time series to all of the subsequences.
Various strategies for determining the most discriminative shapelets have been explored; for example, the binary shapelet transform uses a one-vs-all strategy for multi-class problems \citep{bostrom2015binar}.

\citet{wistuba2015ultra} explores ultra-fast shapelets (UFS), which, in contrast to ST, computes the transformation using a large random sample of subsequences. The study demonstrates that UFS is orders of magnitude faster than ST while enhancing predictive performance. To the best of our knowledge, UFS is among the initial shapelet-based transformation methods that investigate applying first-order differences to augment the available data during the training phase. Nonetheless, it does not distinguish between subsequences derived from the first-order differences and those from the original time series, opting to treat them in the same way.
Similarly, \citet{karlsson2016gener} introduces an ensemble of shapelet-based decision trees constructed using randomly sampled shapelets. This approach improves the speed of inference and the predictive performance compared to UFS and exhaustive shapelet-based decision trees.

Recently, \citet{guillaume2022randp} introduces DST, a new formulation of shapelets that integrates the concept of dilation, effectively expanding the receptive field of the shapelets. The paper explores several innovative ideas, including four methods for enhancing the discriminative power of shapelets. The first method involves using the minimal distance, akin to the approaches of \citet{wistuba2015ultra} and \citet{lines2012shape}, to detect the presence or absence of a subsequence within a given time series. The second method employs the \emph{location} at which the minimal distance occurs to identify shapelets common to both subsequences but differ in their respective locations. The third method utilizes the frequency of occurrence of a shapelet within a given time series, defined as the number of times a shapelet's distance falls below a specified threshold.
This paper demonstrates that these novel methods for constructing a shapelet-based transformation substantially improve performance in subsequent classification tasks. DST represents the current state-of-the-art predictive performance for shapelet-based classifiers.

\subsection{Dictionary-based classification methods}
Dictionary-based transformations represent an approach similar to that of shapelet-based transformations. Both types of transformations utilize phase-independent subsequences. However, in dictionary-based transformations, these subsequences or windows are converted from real values into discrete values, often called words, resulting in sparse feature vectors comprising word counts.

The primary distinction among the various methods for constructing dictionary-based transformations lies in discretizing a window into a discrete word. For instance, BOSS \citep{schafer2015boss}, the temporal dictionary ensemble (TDE) \citep{middlehurst2021tempo}, multi-resolution sequential learner (MrSEQL) \citep{lenguyen2019intera}, and multiple representation sequence miner (MrSQM) \citep{nguyen2023fast} employ symbolic Fourier approximation (SFA). SFA involves $z$-normalizing each subsequence and dimensionality reduction using the initial Fourier coefficients. These coefficients are then discretized into a single symbol within a fixed-size alphabet to form words for counting. The main differences between BOSS, MrSEQL, MrSQM, and TDE reside in how the time series are represented and how the ensembles of representations are formed.

A similar method employing various time series compression techniques, such as symbolic aggregate approximation (SAX), is proposed by \citet{lee2023z-tim} in a dictionary-based transformation termed Z-time. Z-time utilizes temporal abstractions derived from Allen's seven temporal relations (e.g., follows, meets, etc.) and counts their horizontal support.

\subsection{Convolution-based classification methods}

Convolutional-based transformations compute a sliding window over the time series like shapelet-based transformations, while a random vector, not subsequences, represents the kernel. The convolution operators also differ between the two. The convolution operator is the traditional dot product between the kernel and the time series for convolution-based methods. In contrast, the convolution operator is a distance measure for shapelet-based transformations, such as the Euclidean distance.

The first convolution-based transformation, Rocket, is introduced by \citet{dempster2020rocke}. The kernels are randomly dilated, and the time series are either padded such that the center value of the kernel is aligned with the first value of the time series. Rocket convolves the randomly dilated kernels over the time series and computes the global \emph{max pooling} value and the proportion of positive values in the convolution output. Rocket has since been improved to decrease computational cost through MiniRocket \citep{dempster2021minir}, and incorporate it into HIVE-COTE through ensembling \citep{middlehurst2021hive-}

MultiRocket \citep{dempster2021minir} eliminates the stochastic elements and employs a predetermined kernel size, along with a limited set of fixed kernels and dilation levels. Furthermore, MultiRocket computes only the proportion of positive values in the convolution output and utilizes multiple bias values, yielding several features for each kernel. Additionally, MultiRocket \citep{tan2022multi} augments MiniRocket by introducing three additional pooling operations: the mean of positive values, the mean of indices of positive values, and the longest sequence of consecutive positive values. It also enlarges the training set by applying first-order differences to all samples.

More recently, \citet{dempster2023hydra} introduces Hydra, a novel reformulation of the Rocket family of methods, which resembles a dictionary-based method through competing kernels. The convolutional-based algorithms transform time series using numerous random conventional kernels drawn from a normal distribution with zero mean and a standard deviation of one. Hydra \citep{dempster2023hydra} integrates features of Rocket-style algorithms and draws inspiration from dictionary-based methods. Hydra transforms time series with a collection of random convolutional kernels, which, similar to the strategy proposed in this manuscript, are organized into $g$ groups, each containing $k$ kernels. For each time step of the convolutional output within a group, Hydra identifies the kernel with the largest magnitude dot product, the kernel that best matches the input.

\subsection{Other state-of-the-art classification methods}
Over the past ten years, the time series analysis community has achieved considerable advancements in predictive performance. Beyond the methodologies addressed thus far, numerous cutting-edge approaches for time series classification that do not rely on transformation have been introduced, such as TDE, MrSQM, InceptionTime \citep{ismailfawaz2020incep}, TS-CHIEF, and HIVE-COTE 2.0 \citep{middlehurst2021hive-}.  InceptionTime \citep{ismailfawaz2020incep} is currently the most accurate deep learning architecture for time series classification, comprising five ensembles based on the Inception model. Unlike ResNet \citep{he2016deep}, which utilizes residual blocks, Inception-based networks consist of inception modules rather than fully connected convolutional layers. HIVE-COTE 2.0 comprises multiple components, including variants of Rocket called Arsenal and a variant of the time series forest called DrCif.

\section{\castor{}: Competing Dilated Shapelet Transform}\label{sec:methods}
In this section, we introduce the necessary concepts and notations before formally defining our problem. Next, we carefully describe \castor{} (\textbf{C}ompeting Dil\textbf{A}ted \textbf{S}hapelet \textbf{T}ransf\textbf{OR}m) with the following steps: (1) Embedding time series through the utilization of randomly sampled dilated shapelets, (2) Arranging the shapelets into multiple groups wherein they \emph{compete}, and (3) Constructing an efficient transformation of time series, characterized by three features calculated from each shapelet. These features are then applicable for downstream tasks such as classification. Finally, we discuss the hyperparameters and the computational complexity of \castor{}.

\subsection{Preliminaries}\label{sec:background}
A time series is an ordered sequence of measurements, where each measurement has an increasing timestamp. Although time series can be irregularly sampled and have timestamps with missing values, we limit our attention to time series with no \emph{missing values} and are \emph{regularly sampled}.

\smallskip
\begin{definition} {\bf{(time series)}}
	A time series $T=<t_1, t_2, \ldots, t_m>$ is an ordered sequence of $m$ values, where each $t_i \in \mathbb{R}$. We assume $(i \in \mathbb{Z}) \cap (i \in [1,m]) \rightarrow t_i \in T $, i.e., $T$ is regularly sampled. 
\end{definition}

\smallskip
A time series dataset $\mathcal{X}=\{T^j, y^j\}_{j=1}^n$ consists of multiple time series, each labeled with a label $y^j \in \mathcal{Y}$, where $\mathcal{Y}$ denotes a finite set of labels that can be either binary or multi-class.
Given a time series, we can extract patterns, such as \emph{subsequences}, representing characteristics of the time series.

\smallskip
\begin{definition} {\bf{(contiguous subsequence)}}
	A contiguous subsequence $T_{s,l}=<t_s, t_{s+1},\ldots,t_{s+l-1}>$ of a time series $T$ denotes a segment of length $l$ starting at position $s$ in $T$.
\end{definition}
\smallskip

A shapelet $S$ is a special type of subsequence used to calculate \emph{distance profile}. The distance profile is used to capture the discriminative power of a shapelet
$S$, such that the $i^{\text{th}}$ value corresponds to the distance between the shapelet $S$ and the subsequence starting at position $i$ in $T$.

\smallskip
\begin{definition} {\bf{(distance profile)}}
	Given a time series $T$ of length $m$, and a shapelet $S$ of length $l$, the distance profile is given by
	\[
		dp(S, T) = \left[ D(S, T_{i, l}) \mid i \in \{1, \ldots, m - l + 1\} \right],
	\]
	where $D$ is a distance measure, for example, the Euclidean distance as follows:
	\[
		D(S, T_{i,l}) = \sqrt{\sum_{j=1}^{l} (S_j - T_{i+j-1,l})^2}.
	\]
\end{definition}
\smallskip

The distance profile can be regarded as a response analogous to feature maps in \emph{convolutional} neural networks \citep{oshea2015intro}, or the convolution response used by Hydra \citep{dempster2023hydra} and Rocket \citep{dempster2021minir} to form time series transformations.

Dilation expands the kernel by inserting empty spaces between kernel values. We can dilate subsequences by a factor $d$ by inserting $d-1$ empty spaces between values. Setting $d=1$ corresponds to a non-dilated subsequence.

\smallskip
\begin{definition} {\bf{(dilated subsequence)}}
	A subsequence of length $l$ with dilation $d$ and offset $s$, where 
	$s + (l-1) \cdot d \leq m$ is obtained by repeatedly incrementing the offset by the dilation rate, i.e., $T_{d,s,l} = <t_{s}, t_{s+d}, t_{s+2d}, \ldots, t_{s + (l - 1)d}>$. A subsequence of length $l$ dilated by a factor $d$ has an effective length $\hat{l}=(l-1)d+1$.
\end{definition}
\smallskip

By dilating the subsequence and introducing gaps, we can expand the receptive field of the subsequence, similar to the effects of downsampling. Consequently, a dilated distance profile can reveal features across different scales of granularity \citep{oord2016waven}.

\begin{figure}
	\begin{center}
		\includegraphics[width=1.0\textwidth]{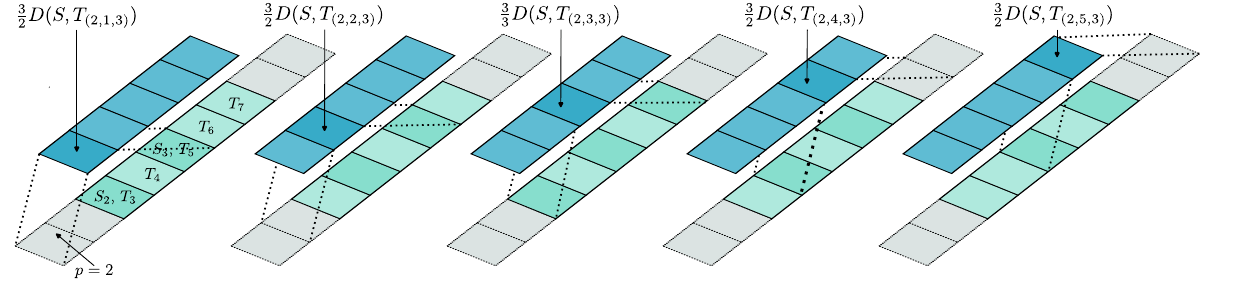}
	\end{center}
	\caption{Example of the dilated distance profile (blue) of a shapelet $S$ of length $3$ and a time series $T$ of length $5$ with dilation $d=2$ and padding $p=2$. We slide the shapelet over the time series, and at each timestep $i$, we compute the distance $D$ between the shapelet and the matching positions in the time series. Note that we scale the distance according to the number of timesteps that are not part of the padding, effectively assuming that the distance inside the padding is equal to the distance inside the time series.}\label{fig:dilated-distance-profile}
\end{figure}

\smallskip
\begin{definition} {\bf{(dilated distance profile)}}
	Given a time series $T$ of length $m$ and a subsequence $S$ of length $l$ with dilation $d$, we define the dilated distance profile as follows:
	\[
		dp_d(S, T) = \left[ D(S, T_{d, s, l}) \mid s \in \{1, \ldots, m - (l - 1)d + 1\} \right],
	\]
	with $dp_1(S, T) = dp(S, T)$. The output size of the dilated distance profile is given by $o = m - (l - 1)d+1$.
	\label{def:dilated-distance-profile}
\end{definition}
\smallskip

The size of the distance profile diminishes in relation to the \emph{effective size} of the subsequence $T_{d, s, l}$. To regulate the size of the distance profile, we pad the input time series $T$ with empty values as follows:

\smallskip
\begin{definition} {\bf{(padded time series)}}
	Given a time series $T$ of length $m$, a padded time series $\hat{T}=<0_1, \ldots, 0_p, t_1, \ldots, t_m, 0_1, \ldots, 0_p >$ with padding $p$ has an effective length of $\hat{m} = m + 2p$.
	\label{def:padded-time-series}
\end{definition}
\smallskip
By combining Definition~\ref{def:padded-time-series} with Definition~\ref{def:dilated-distance-profile}, we define the padded and dilated distance profile in which we pad the time series $T$ and dilate the shapelet $S$. Henceforth, we regard \emph{distance profile} as being padded and dilated by default unless explicitly stated otherwise.

\smallskip
\begin{definition} {\bf{(padded dilated distance profile)}}
	Given a time series $T$ of length $m$ padded with $p$ empty values and a subsequence $S$ with dilation $d$, the padded dilated distance profile is defined as:

	\[
		dp_{d, p}(S, T) = \left[ D(S, \hat{T}_{d, s, l}) \mid s \in \{1, \ldots, m+2p - (l - 1)d+1\}\right].
	\]
	The output size of the padded dilated distance profile is given by $o=m+2p-(l-1)d+1$.
\end{definition}
\smallskip

To maintain an output size equivalent to the input size ($o=m$), we define $p=\left\lfloor\frac{(l-1)d+1}{2}\right\rfloor$\footnote{For simplification, we ensure that $l$ is odd to avoid asymmetric padding.}. Note that we do not consider the padded empty values in $T$ for the distance calculation. When the distance calculation includes padded empty values, we scale up the distance by the number of non-empty values to make the scales of distances equivalent regardless of the padding. Assuming $\hat{T}_{d,s,l}$ is a corresponding subsequence of the shapelet $S$, We calculate the distance as follows:

\[
	\frac{l}{|\{1 \mid j \in \{s, s+d, \dots, s+d \cdot l\},  p < j \leq m + p \}|}D(S, \hat{T}_{d,s,l}),
\]

\noindent where $l$ in numerator is the length of the shapelet $S$, and the denominator is the number of values of $T_{d,s,l}$ outside of the padded area being considered. 

Figure~\ref{fig:dilated-distance-profile} shows an example of computing the dilated distance profile of a shapelet of length $3$ on a time series of length $5$ with $d=2$ and $p=2$, resulting in a distance profile of size $5$. Given these preliminaries, we are now in a position to explain the inner workings of the \castor{} transformation.

\subsection{\castor{}: An overview}\label{sec:overview}
\castor{} is a novel shapelet-based time series transform that uses dilation and incorporates the properties from both dictionary-based and convolution-based transforms. In \castor{}, time series time series undergo transformation through a set of dilated shapelets, organized into $g$ groups. At each time step of the distance profile, produced by sliding dilated shapelets across the padded time series, we extract three features per shapelet. These features encapsulate the efficacy of each shapelet within the $g$ groups in representing subsequences, as measured by distance metrics such as smallest/largest distances and distances within a specified threshold.

By randomly sampling shapelets, \castor{} can explore the data and discover discriminatory patterns within the time series, similar to \citet{wistuba2015ultra,guillaume2022randp}, while avoiding the exploration of all possible shapelets, which requires significant computational resources. In contrast, Hydra and Rocket explore completely random patterns drawn from a normal distribution \citep{dempster2021minir,dempster2023hydra}, essentially ignoring discriminatory information present in the training data.

In contrast to DST \citep{guillaume2022randp}, which also computes features based on the distance profile, \castor{} arranges multiple shapelets into $g$ groups to allow them to \emph{compete} to construct the transforms. Similar to Hydra, MultiRocket, and DST, \castor{} uses exponential dilation, i.e., the dilation parameter is set to $d=2^e$ where $e\in\left[0, E\right]$ with $E=\lfloor\text{log}_2 \frac{m}{l}\rfloor + 1$. Instead of drawing a random dilation for every subsequence as \citet{guillaume2022randp}, we randomly sample $k$ shapelets for all dilation levels for each level where $k$ shapelets are dilated with the same factor $2^e$ where $e \in E$. Thus, we create $E$ groups where each group contains $k$ shapelets for each group, creating in total $E \times k$ shapelets. We denote a group of shapelets with exponential dilation as $\mathbf{G}^k_d$.

\smallskip
\begin{definition} {\bf{(dilation group)}}
	A dilation group of $k$ shapelets with dilation $d$ is defined as $\mathbf{G}^k_d = \{S_1, \ldots, S_k\}$.
\end{definition}
\smallskip

We randomly sample $k$ shapelets from a set of time series with a specified length $l$ to form the group as
	\[
	\mathbf{G}^k_d = \{T^{j}_{d, s, l}\}^{k},
	\]
\noindent where $j$ is randomly drawn from a discrete uniform distribution over the set $\{1, n\}$ and $s$  is randomly drawn from a discrete uniform distribution over the set $\{1, m - \hat{l}\}$.

The final \castor{} parameters comprise $g \times E \times k$ shapelets, with each group containing $k$ shapelets for each dilation level.
Since $E$ grows logarithmically with the time series length, the number of output features depends on the input size.


\smallskip
\begin{definition}  {\bf{(\castor{} parameters)}}
	The \castor{} parameters, consisting of $g$ groups with $k$ randomly sampled shapelets and exponential dilation, are defined as
	\[
		\mathcal{G} = \left\{\{ \mathbf{G}^k_d \mid d \in \{2^0, \ldots, 2^E\} \}\right\}^g.
	\]
\end{definition}
\smallskip
\begin{figure}
	\begin{center}
		\includegraphics[width=0.8\textwidth]{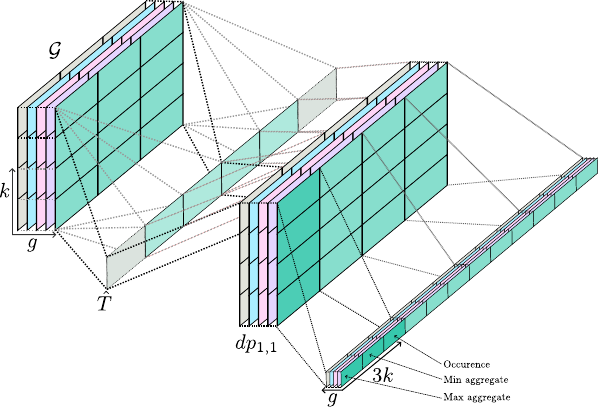}
	\end{center}
	\caption{A simplified representation of \castor{}. We have \castor{} parameters of size $g \times E \times k \times l = 5 \times 1 \times 4 \times 3$ computing the distance profile for every subsequence in $\hat{T}$ of length $m=4$ with padding $p=1$ and dilation $d=1$, resulting in the distance profiles of shape $5 \times 4 \times 4$. Inside each group and for a given column in the distance profiles, we compute three distance-based features ($3k$ in total as shown on the right side, see Section ref) for each shapelet based on the minimum and maximum distance and the distance threshold. Finally, we flatten the array.\label{fig:embedding}}
\end{figure}

Given that the dimensions of all components in $\mathcal{G}$ are known with $g$ representing the number of groups, $k$ denoting the number of shapelets, $E$ indicating the number of exponents, and $l$ signifying the shapelet length, it is possible to conceptualize $\mathcal{G}$ as a multidimensional array with the dimensions $g \times k \times E \times l$. Figure~\ref{fig:embedding} illustrates the computation of the transformation for a single padded time series $\hat{T}$ with length $m=4$, utilizing five groups, each comprising four shapelets of length $l=3$ and a single dilation factor $d=0$, which means $E$ comprises a single exponent $e=1$. For each group, the distance profile between every shapelet in all groups and $T$ is computed, yielding $5 \times 4$ distance profiles of length $o=4$. For each column corresponding to a time step in the distance profile, features are computed based on the distance profile values for each shapelet within the group.

We present pseudocode for training the \castor{} representation in Algorithm~\ref{alg:fit_castor}, executing the transformation in Algorithm~\ref{alg:transform_castor}, and provide an efficient implementation of \castor{} with interface to Python along with the main competitors: Rocket, Hydra, DST, and UST in Wildboar \citep{samsten2024isaks}. The implementations are compatible with \texttt{scikit-learn} \citep{pedregosa2011sciki}, from which we also use the Ridge classifier to evaluate the transformation. The complete source code is available on GitHub\footnote{\url{https://github.com/wildboar-foundation/wildboar}} and the experiments, as well as all data, are in the supporting repository\footnote{\url{https://github.com/isaksamsten/castor}}.


\begin{algorithm}
	\Input{%
		\begin{tabular}[t]{ll}
			$\mathcal{X}$ & Training set  \\
			$g$           & No. of groups    \\
			$k$           & No. of shapelets \\
			$l$           & Shapelet length
		\end{tabular}\qquad
		\begin{tabular}[t]{ll}
			$\rho_{lower}$  & Lower occurrence             \\
			$\rho_{upper}$ & Upper occurrence             \\
			$\rho_{norm}$ & Z-normalization probability \\
		\end{tabular}
	}
	\Output{%
		\begin{tabular}[t]{ll}
			$\mathcal{G}$ & \castor{} parameters of shape $g \times k \times E \times l$ \\
			$\Lambda$     & Occurrence thresholds of shape $g \times k \times E$         \\
		\end{tabular}
	}
	\BlankLine{}
	$E \gets \lfloor \log_2 \frac{m}{l}\rfloor + 1$  \Comment*[r]{Maximum number of dilations.}
	$\mathcal{G} \gets$ array of shape $(g, k, E, l)$ \;
	$\Lambda \gets$ array of shape $(g, k, E)$ \;
	\For{$i \gets \{1, \ldots, g\}$} {
		\For{$j \gets \{1, \ldots, k\}$} {
			\For{$e \gets \{1, \ldots, E\}$} {
				$d \gets 2^e$, $\hat{l} \gets (l - 1) d + 1$, $p=\frac{\hat{l}}{2}$ \Comment*[r]{Padding and dilation}
                $a \gets \mathcal{U}_{[1, n]}$ \Comment*[r]{Sample shapelet}
                $b \gets \mathcal{U}_{[1, n]}$ such that $a\neq b$ and $y^a = y^{b}$\;
                $\mathcal{G}_{i, j, e} \gets T^{a}_{d, \mathcal{U}_{[1, m - \hat{l}]}, l}$ \;
				$z$-normalize $\mathcal{G}_{i, j, e}$ with probability $p_{norm}$ \;
				$dp_{d, p}(\mathcal{G}_{i, j, e}, T^{b})$ is calculated and sorted in ascending order \;
                $\Lambda_{i, j, e} \gets  dp_{d,p}(\mathcal{G}_{i, j, e}, T^{b})_{\mathcal{U}_{[\lfloor \rho_{lower} \cdot m\rfloor, \lfloor \rho_{upper} \cdot m\rfloor]}} $ \;

   }
		}
	}
	\caption{Creation of the \castor{} parameters.}\label{alg:fit_castor}
\end{algorithm}

Algorithm~\ref{alg:fit_castor} illustrates the construction of the \castor{} parameters.
The algorithm begins by initializing two multidimensional arrays, $\mathcal{G}$ and $\Lambda$, which contain the dilated subsequences and the distance thresholds for computing occurrences, respectively. For each group $\{1, \ldots, g\}$, shapelet $\{1, \ldots, k\}$, and exponent $\{0, \ldots, E\}$, \castor{} samples shapelets from a random time series in $T$ with a random start position and the specified length $l$. Then
a sampled subsequence is $z$-normalized with probability $\rho_{norm}$. Subsequently, the algorithm computes the dilated distance profile between the shapelet extracted from $T^a$ and another randomly selected time series $T^{b}$, ensuring that $a \neq b$ and that both $T^a$ and $T^{b}$ share the same class label. In the final step, the occurrence threshold is determined by sampling from the $\rho_{lower} \cdot m$ to $\rho_{upper} \cdot m$ smallest distances in the distance profile, returning the \castor{} parameters and the occurrence thresholds for every shapelet.



\begin{algorithm}
	\Input{ %
		\begin{tabular}[t]{ll}
			$\mathcal{G}$        & \castor{} representation \\
			$\Lambda$            & Occurrence thresholds      \\
			$\{T^1,\ldots,T^n\}$ & Time series
		\end{tabular}
	}
	\Output{$\mathbf{O}$ : array of shape $(n, g * k * E * 3)$}
	\BlankLine{}
	Set $g, k, E, l$ from the shape of $\mathcal{G}$\;
	$\mathbf{O} \gets$ array of shape $(n, g, E, k * 3)$ \Comment*[r]{Three features}
	\For{$a \gets \{1, \ldots, n\}$}{
		\For{$i \gets \{1, \ldots, g\}$}{
			\For{$e \gets \{1, \ldots, E\}$}{
				$d \gets 2^e$, $\hat{l} \gets (l - 1) * d + 1$, $p=\frac{\hat{l}}{2}$ \Comment*[r]{Padding and dilated size}
				$\mathbf{dp} \gets $ array of shape $(k, m)$ \;
				\For{$j \gets \{1, \ldots, k\}$}{
					$\mathbf{dp}_{j} \gets dp_{d,p}(\mathcal{G}_{i,j,e}, T^{a})$ \;
				}
				\For{$j \gets \{1, \ldots, k\}$}{
                    $\mathbf{O}_{a, i, e, j} \gets MinAggregate(j, \mathbf{dp}, \text{soft})$ \Comment*[r]{See Eq. 1}
                    $\mathbf{O}_{a, i, e, k+j} \gets MaxAggregate(j, \mathbf{dp}, \text{hard})$ \Comment*[r]{See Eq. 2}
                    $\mathbf{O}_{a, i, e, 2k+j} \gets Occurrence(j, \mathbf{dp}, \Lambda_{i, j, e})$ \Comment*[r]{See Eq. 3}
                }    
			}
		}
	}

	Reshape $\mathbf{O}$ to shape $(n, g * E * k * 3)$ \Comment*[r]{Flatten to feature vectors}
	\caption{Calculating transforms using the \castor{} parameters.}\label{alg:transform_castor}
\end{algorithm}

Algorithm~\ref{alg:transform_castor} illustrates the process for transforming time series data using the \castor{} parameters. For each sample index in the set $\{1, \ldots, n\}$, each group index in the set $\{1, \ldots, g\}$, and each exponent in the set $\{0, \ldots, E\}$, the algorithm initializes a $k\times m$ matrix to store the distance profiles. Subsequently, the algorithm calculates the distance profile between each shapelet, indexed by $\{1, \ldots, k\}$, and the time series. In the final step, from each shapelet, the algorithm determines the following three features to use: (1) the aggregation of minimal distances, (2) the aggregation of maximal distances, and (3) the occurrences for all subsequences above the threshold. We discuss how we compute these features in Section~\ref{sec:subsequence-features}. 

After all samples have been transformed, the transformation matrix is flattened to the feature vectors for time series. The transformed time series can subsequently be used for classification, e.g., using the traditional ridge classifier popularized by \citet{dempster2020rocke}.


\subsection{Subsequence features}\label{sec:subsequence-features}
Give a shapelet $S$ and two (or more) time series $T^a$ and $T^b$, there are several options for discriminating between the two using the distance profiles.
In particular, one can consider:

\begin{itemize}
	\item The shapelet $S$ exhibits greater similarity to time series $T^a$ than to time series $T^b$, indicating that the shapelet is present in $T^a$ but absent in $T^b$. 
	\item The shapelet $S$ is absent in both $T^a$ and $T^b$; in other words, the shapelet does not exhibit similarity to either time series. This implies the presence of an anomalous pattern, suggesting that the shapelet demonstrates distinct dynamics, is associated with a different label, or captures noise. 
	\item The shapelet $S$ may be present in both time series $T^a$ and $T^b$ but at different location.
	\item The shapelet $S$ may be present in both time series $T^a$ and $T^b$, yet it occurs more frequently in one of them, signifying that the shapelet forms a recurrent pattern. 
	\item The shapelet $S$ manifests an identical morphology in time series $T^a$ and $T^b$, albeit at different scales, suggesting that the \emph{pattern} is distinctive, yet it presents with varying magnitudes.
\end{itemize}
\castor{} captures the aforementioned properties by the three feature values calculated for each shapelet in one group $\mathbf{G}^k_d$: \emph{minimum distance}, \emph{maximum distance}, and \emph{frequency}. First, the count of minimal distances for each time series at each time point quantifies the similarity between the shapelet in the group and the corresponding location.
Second, the count of maximal distances measures the degree of dissimilarity between a shapelet within the group and the corresponding location. Both features detect the presence and absence of a shapelet, as well as the specific locations where it occurs or is missing. Finally, the frequency of a shapelet within a group signifies the occurrence of repetitive patterns in the time series. By analyzing the frequency, it is possible to identify \emph{templates} or \emph{motifs} that characterize the recurring features or patterns in the behavior of the time series.

\subsubsection{Competing features}

We aim to find the most representative features out of the shapelets within each group $\mathbf{G}^k_d$, each characterized by the same dilation factor $d \in \{1, \dots, E\}$. By allowing them to \emph{compete} at each discrete time step of the time series, \castor{} enables a more accurate representation of explicit temporal behavior.
Competing means that we generate features based on how well the shapelets in the same group represent each discrete time step of one time series in terms of relative similarity or dissimilarity to subsequences. 
If a shapelet $S_i$ has the minimum distance among the shapelets in $\mathbf{G}^k_d$ at time step $j$, meaning it has the minimum value at the $j$-th index of the distance profile, we increment its frequency by one (if $\delta = \text{hard}$) or its maximum or minimum value (if $\delta = \text{soft}$).
Given the distance profiles of all subsequences in a group $i$ and dilation $d$, denoted as $\mathbf{dp} = \left\{ dp_{d,p}(\mathcal{G}_{i,j,e}, T) \mid j \in \{1, \ldots, k\}\right\}$, which form a matrix of dimensions $k \times m$, the frequency with which a shapelet $S_i$ attains the minimum distance at each time step of $T$ is quantified as follows:

\begin{equation}
	\text{MinAggregate}(j, \textbf{dp}, \delta) = \sum_{i=1}^m \mathbf{1}(\argmin \textbf{dp}_{:,i} = j) \cdot
 \begin{cases}
   {1}, & \text{if } \delta = \text{hard}\\
     \min\textbf{dp}_{:,i},              & \text{otherwise}
\end{cases}
\label{eq:hard-min}
\end{equation}

We additionally consider the converse scenario, where we determine the frequency with which a subsequence within a group exhibits the maximal distance at time step $j$. This calculation can be performed by employing Equation~\ref{eq:hard-min} and replacing $\argmin$ with $\argmax$ (i.e., \emph{hard} counting),

\begin{equation}
	\text{MaxAggregate}(j, \textbf{dp}, \delta) = \sum_{i=1}^m \mathbf{1}(\argmax \textbf{dp}_{:,i} = j) \cdot
 \begin{cases}
   {1}, & \text{if } \delta = \text{hard}\\
     \max\textbf{dp}_{:,i},              & \text{otherwise}
\end{cases}
	\label{eq:hard-max}
\end{equation}

As such, hard counting involves performing an $\argmax$ or $\argmin$ operation over the distance profiles for each group, yielding the index of the subsequence with the maximum or minimum distance at each time step. Subsequently, the index corresponding to the minimal or maximal subsequence is incremented. While the maximal and minimal distances capture distinct discriminatory characteristics, high minimal counts should be observed when the subsequence is positively discriminatory, whereas high maximal counts suggest negative discrimination.
Similar to Hydra, we also explore \emph{soft} counting of the minimal and maximal distance, essentially replacing adding ones with adding actual $\min$ and $\max$ values when the $\argmax$ or $\argmin$ condition suffices in Equation~\ref{eq:hard-min} and Equation~\ref{eq:hard-max} respectively.

We consider the selection of the counting strategy as a hyperparameter and permit any combination of strategies. In Section~\ref{sec:ablation}, we demonstrate that a hybrid approach, utilizing soft counting for the minimal distance and hard counting for the maximal distance, results in transformations with enhanced predictive performance.
Consequently, the standard configuration for \castor{} is to utilize soft counting for the minimum distance and hard counting for the maximum distance.

\subsubsection{Shapelet occurrences}
Besides the two features extracted from competing shapelets, \castor{} also considers the third feature derived from the \castor{} parameters (i.e., $\mathcal{G}$), namely the count of shapelet occurrences.
Given a vector of $k$ distance thresholds $\lambda$ with each threshold associated with a shapelet, we can capture the \emph{independent} occurrence of the shapelet as follows:

\begin{equation}
	\text{Occurence}(j, \textbf{dp}, \lambda) = \sum_{i=1}^m \textbf{1}(\mathbf{dp}_{i,j} < \lambda_j),
	\label{eq:occurence}
\end{equation}
which, given a distance profile, quantifies the number of time steps for which the $i$-th shapelet remains below a specified threshold.
For a given shapelet, the threshold to signify occurrence can be determined through several methods, such as calculating the distance to a randomly chosen shapelet of a time series, or by selecting a threshold at random from a predetermined range.

Akin to the approach described by \citet{guillaume2022randp}, \castor{} calculates the threshold in the following manner. Given a shapelet S extracted from a time series $T^1$, we determine the distance profile between $S$ and another time series $T^{2}$, denoted as $dp_{d,p}(S, T^{2})$, where $T^{2}$ shares the same label as $T^{1}$. Subsequently, we arrange the distance profile in ascending order and uniformly select a threshold value from within the range delineated by two percentiles, $[\rho_{lower}, \rho_{upper}]$. The process is formalized as follows:

\begin{equation}
	\lambda = \left\{ dp_{d,p}(S_j, \hat{T}^2)_{i} \mid j \in \{1, \ldots, k\},\ i \sim \mathcal{U}_{[\lfloor \rho_{lower} \cdot n\rfloor, \lfloor \rho_{upper} \cdot n\rfloor]} \right\},
	\label{eq:lambda}
\end{equation}
where we assume that $dp(\cdot, \cdot)$ returns values sorted in ascending order. We calculate Equation~\ref{eq:lambda} for all shapelet groups $\mathcal{G}$ as shown in Algorithm~\ref{alg:fit_castor}, creating the matrix $\Lambda$.


To increase the diversity of the embedding and reduce the computational complexity, our strategy involves randomness rather than selecting the optimal occurrence threshold, similar to \citet{guillaume2022randp}. In Section~\ref{sec:ablation}, we examine the trade-offs of the selections of these two percentiles.

Similar to the distinction between hard and soft counting of the minimal and maximal subsequence distances, we can also differentiate between \emph{independent occurrence} and \emph{competitive occurrence}. In the case of competitive occurrence, we calculate the frequency at which the subsequence exhibiting the minimal distance in a group falls below the threshold. In contrast, in independent occurrence, we calculate the frequency at which any subsequence falls below the threshold.
Since we demonstrate that employing independent occurrence yields greater accuracy than utilizing competitive occurrence (see Section~\ref{sec:ablation}), \castor{} employs independent occurrence by default.

\subsubsection{Subsequence normalization}
Scaling the distance between shapelets has been crucial for many time series classifiers and transforms \citep{ye2009time}.
\castor{} incorporates this aspect by computing the distance profile between shapelets and time series under $z$-normalization.


We incorporate a parameter $\rho_{\text{norm}}$ in Algorithm~\ref{alg:fit_castor}. This parameter governs the probability with which a group $\mathbf{G}^k_{2^0,\ldots,2^E}$ utilizes normalized or non-normalized distance profiles. In Section~\ref{sec:ablation}, we demonstrate that neither of the two extremes—consistently normalizing or consistently not normalizing—yields transformations with superior accuracy. Consequently, we set $p_{\text{norm}}=0.5$ by default, which means that we use normalized distance profiles for half of the groups.

\subsection{First-order difference}
Similar to many state-of-the-art time series classifiers and embeddings \citep[see, e.g.,][]{tan2022multi,dempster2023hydra,guillaume2022randp,middlehurst2020canonical}, \castor{} processes both the original time series and its first-order difference. For a given time series, the first-order difference is defined as:
\[
	T'=\{T_i-T_{i-1} \mid i \in \{2, \ldots, m\}\}.
\]
By default, the first-order difference of the time series is pre-computed.
Following Hydra, time series only in half of the groups (i.e., $g/2$ groups) are transformed, and the remaining $g/2$ groups keep the original time series.
This approach enables the embedding to capture salient features in both the original and the differenced representations. We execute Algorithm~\ref{alg:fit_castor} on both time series $T$ and $T'$, with the parameter $g$ set to $g/2$ for each, yielding the \castor{} parameters $\mathcal{G}$ and $\mathcal{G}'$ corresponding to the original and differenced time series, respectively. Subsequently, we employ Algorithm~\ref{alg:transform_castor} on $T$ utilizing $\mathcal{G}$ and on $T$ utilizing $\mathcal{G}'$. The resulting feature vectors are then concatenated to construct the final transformation.

By incorporating first-order differences, we enhance the heterogeneity of the training set, akin to the approach described in \citet{middlehurst2021hive-}. This strategy enables us to construct an ensemble of transformations across various input representations. Specifically, first-order differences quantify the rate of change between successive time points within a time series.
This rate of change constitutes one of numerous potential representations that have been demonstrated to improve predictive performance \citep{middlehurst2021hive-,dempster2023hydra}. In accordance with the findings of \citet{dempster2023hydra}, we report in Section~\ref{sec:ablation} that the use of first-order differences enhances the predictive accuracy of \castor{} on the development datasets. Consequently, we adopt this technique as one of the default hyperparameters.

\subsection{Computational complexity}
The computational complexity of \castor{} is influenced by the parameters $g$, $k$, the number of samples $n$, and the number of time steps $m$.
The predominant computational expense is attributed to the calculation of the distance profile, which, for a shapelet of length $l$, is $\mathcal{O}(m\cdot l)$.
Consequently, for fitting the transformation, we compute $g \times \log_2(m) \times k$ distance profiles, resulting in a total computational cost of $\mathcal{O}(g \times k \times \log_2(m) \cdot m \cdot l)$. However, since $g$ and $k$ are constants, they do not affect the asymptotic complexity, which simplifies to $\mathcal{O}(\log_2(m) \cdot m \cdot l)$.

Similarly, for applying the transformation with already fitted \castor{} parameters, we calculate $g \times \log_2(m) \times k$ features for each of the $n$ samples.
Next, the feature calculation process for each group involves comparing all $k$ shapelets across the $m$ values in the distance profile, introducing an additional computational term of $\mathcal{O}(m \times k)$.
Therefore, the final computational cost for transforming a collection of time series is $\mathcal{O}(n \cdot g \cdot k \cdot \log_2(m) \cdot m \cdot l + m \cdot k)$, which simplifies to $\mathcal{O}(n \cdot \log_2(m) \cdot m \cdot l)$ since $g$ and $k$ are constants. We experimentally confirm the asymptotic analysis of the computational cost in Section~\ref{sec:ablation}.

\section{Experiments}\label{sec:results}
In this section, we assess the performance of \castor{} on the datasets in the UCR univariate time series archive \citep{bagnall2018uea}. We demonstrate that \castor{} outperforms current state-of-the-art methods in terms of accuracy, including Rocket \citep{dempster2020rocke}, MultiRocket \citep{tan2022multi}, Hydra \citep{dempster2023hydra}, and recent shapelet-based approaches such as RDS \citep{guillaume2022randp} and UST \citep{wistuba2015ultra}. Furthermore, we conduct a comprehensive ablation study to investigate the influence of \castor{}'s hyperparameters on its predictive accuracy and investigate how the number of samples and number of time steps affect the computational cost of the transformation.

\subsection{Experimental setup}

The experimental evaluation was performed on a comprehensive collection of 112 datasets from the UCR archive. While many recent studies in time series classification use pre-defined train-test split of those datasets, we employ 5-fold cross-validation repeated five times for each dataset to evaluate different possible randomness both in the datasets and the algorithms. Consequently, the reported results represent the mean of 25 separate applications of the algorithm. Moreover, we utilize our own highly optimized versions of all comparative methods, except for MultiRocket and DrCif, for which we employ the implementations provided by the original authors.

To evaluate the predictive performance, we configured the parameters of \castor{} based on the findings from the development set (see Section~\ref{sec:ablation})  as follows: $g=128$, $k=16$, $\rho_{\text{lower}}=0.01$, $\rho_{\text{upper}}=0.2$, $\rho_{\text{norm}}=0.5$, using first-order differences for half the groups.
We also employed the \emph{soft minimum} and \emph{hard maximum}, along with independent occurrences for feature representations. For the other methods, we applied the default parameters recommended in the original publications. An exception was made for Hydra, for which we set the number of groups to $g=64$ and the number of kernels to $k=32$. This modification guarantees a fair comparison between \castor{} and Hydra to make Hydra employ the same number of effective parameters as \castor{} (i.e., $2048$). To further guarantee a fair comparison, We also tested both \castor{} and Hydra under Hydra's default parameters ($g=64$ and $k=8$).

For \castor{}, MultiRocket, Hydra, and Rocket, we utilize the standard leave-one-out cross-validated Ridge classifier, as recommended by the respective authors. We employ the implementation from \texttt{scikit-learn} with the default parameters as of version 1.3, which optimizes the regularization parameter $\alpha$ over $\{0.01, 1.0, 10\}$. Furthermore, for MultiRocket and Rocket, the transformed feature values are standardized to have a zero mean and a standard deviation of one. For Hydra and \castor{}, we adopt the feature normalization method proposed by \citet{dempster2023hydra}, designed to accommodate the high prevalence of zeros in the transformed data.

\begin{figure}
	\begin{center}
		\includegraphics[]{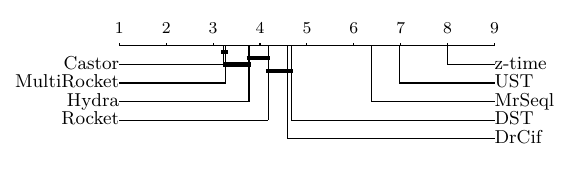}
	\end{center}
	\caption{Mean rank of \castor{} for 5-fold cross-validation, repeated five times across 112 datasets from the UCR time series repository. A lower rank indicates higher accuracy. \castor{} is significantly more accurate ($p<0.01$) than current state-of-the-art methods such as Hydra and Rocket, and is slightly better ranked than MultiRocket.}\label{fig:ranks}
\end{figure}

\subsection{Empirical investigation}
\begin{figure}
	\begin{center}
		\includegraphics[width=\textwidth]{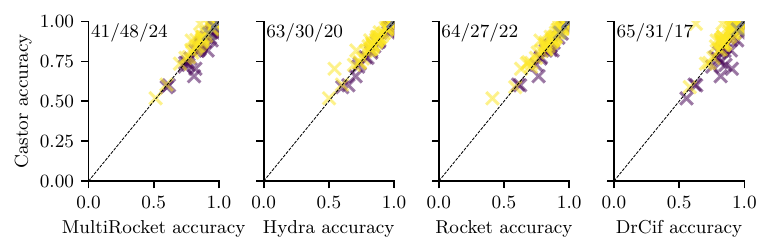}
	\end{center}
	\caption{A \emph{sharpshooter} plot comparing the performance of \castor{} against MultiRocket, Hydra, Rocket, and DST. Annotations in the left corners indicate the wins, losses, and ties for \castor{}, respectively.}\label{fig:sharpshooter}
\end{figure}

Figure~\ref{fig:ranks} presents the mean rank of \castor{} compared to state-of-the-art methods such as MultiRocket, Rocket, Hydra, DTS, interpretable classifiers such as MrSeql and z-time, and the baseline method UST. The results demonstrate that \castor{} is significantly more accurate ($p<0.01$) than all the competing methods, with the exception of MultiRocket. For MultiRocket, the results suggest that \castor{} has a marginally better rank. A sharpshooter plot, depicted in Figure~\ref{fig:sharpshooter}, reveals that although \castor{} has an average rank slightly better than that of MultiRocket, MultiRocket exhibits better accuracy for 50 of the datasets, as opposed to \castor{}, which outperforms MultiRocket in terms of accuracy for 43 datasets.
The discrepancy between the two results may be attributed to the high correlation in the performance of Rocket, Hydra, and MultiRocket, which leaves the determination of the superior algorithm among these methods unresolved, and \castor{} as the best ranked method on average. In fact, the significance tests reveals that there is no statistically significant difference in performance between Hydra and MultiRocket or between Rocket and Hydra, whereas \castor{} has a significant difference in performance compared to both Hydra and Rocket. This confirms that \castor{} is the best shapelet-based time series classifier in terms of predictive performance outperforming the majority of its state-of-the-art counterparts.

\begin{figure}
	\begin{center}
		\includegraphics[width=\textwidth]{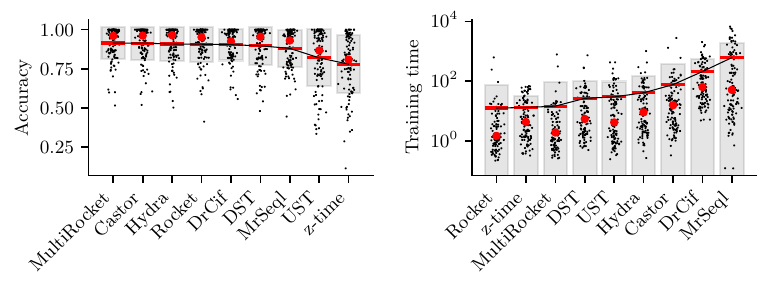}
	\end{center}
	\caption{Accuracy (left) and log-scaled training time (right) for all compared methods across all 112 datasets. Each black dot represents the accuracy for a method over a dataset, the red vertical bar indicates the mean, the red circle denotes the median, and the gray box illustrates the standard deviation.}\label{fig:mean-accuracy-fit-time}
\end{figure}

\begin{figure}
	\begin{center}
		\includegraphics[width=\textwidth]{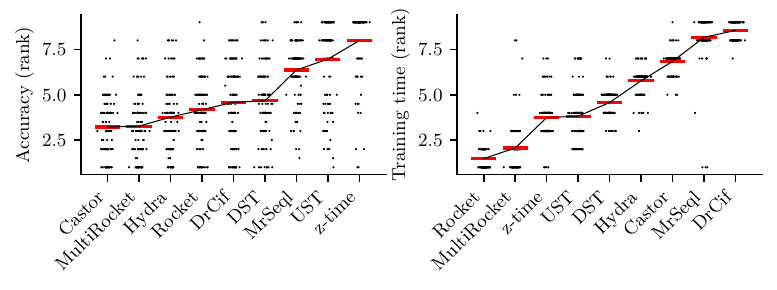}
	\end{center}
	\caption{Average rank for accuracy (left) and training time (right) for all compared methods across all 112 datasets. Each black dot represents the rank for a method over a dataset and the red vertical bar illustrates the mean rank.}\label{fig:rank-accuracy-training-time}
\end{figure}

We examine the rankings of all methods in Figures~\ref{fig:mean-accuracy-fit-time} and \ref{fig:rank-accuracy-training-time}, confirming that the predictive performance of \castor{} and MultiRocket is nearly identical, in terms of both ranking and average performance. Additionally, the standard deviation and median performance of both methods are also nearly equivalent. Furthermore, we observe that Hydra and Rocket, as well as Rocket and DST, form two distinct performance clusters, with no significant difference between Hydra and Rocket, and no significant difference between Rocket and DST.
In terms of computational cost, an inverse relationship is apparent, with Rocket exhibiting the lowest computational cost, both in terms of rank and runtime,  whereas \castor{} is slower than the Rocket variants but faster than DrCif and MrSeql.

\begin{figure}
	\begin{center}
		\includegraphics[width=\textwidth]{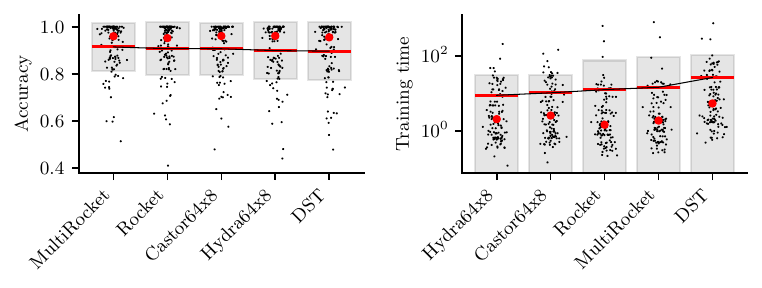}
	\end{center}
	\caption{Average accuracy (left) and training time (right) for all compared methods across all 112 datasets for \castor{}, Hydra, Rocket and MultiRocket, with \castor{} and Hydra fit with $g=64$ and $k=8$ which are suggested as default parameters for Hydra by \citet{dempster2023hydra}}\label{fig:runtime-default-parameters-hydra}
\end{figure}

In Figure \ref{fig:mean-accuracy-fit-time}, we present the accuracy and training time achieved using our default parameter setting. Additionally, in Figure \ref{fig:runtime-default-parameters-hydra}, we compare the accuracy and computational times of the main competitors (i.e., Rocket, MultiRocket, Hydra) with \castor{}. For this comparison, we set the parameters for both \castor{} and Hydra to the default values suggested by \citet{dempster2023hydra} ($g=64$ and $k=8$). While we are in favor of Hydra, the results indicate that both \castor{} and Hydra exhibit lower runtime values than Rocket and MultiRocket under this parameter configuration, albeit with a slight decrease in predictive performance. This confirms \castor{}'s ability to accelerate computations compared to the Rocket variants, particularly under lighter parameter settings, while still maintaining predictive performance levels similar to those achieved under default settings.

\begin{figure}
	\begin{center}
		\includegraphics[width=\textwidth]{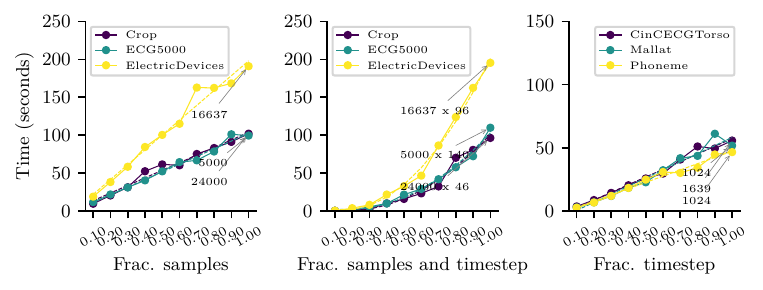}
	\end{center}
	\caption{Runtime for \castor{} for a selection of datasets, scaling the number of samples and the number of time step according to the factor on the x-axis. The total number of samples (left) and total number of time steps (right) of each dataset is annotated in the figures. We can observe through the dashed linear fit that \castor{} scales linearly with both the number of samples and number of time steps.}\label{fig:scaling}
\end{figure}

In Figure \ref{fig:scaling}, we investigate the scalability of \castor{} with respect to both an increasing number of samples (left) and an increasing number of time steps within the time series (right)\footnote{The runtime is computed using all cores on a MacBook Pro M2 with 10 cores}.
The analysis utilizes six datasets that exhibit specific characteristics; \emph{Crop}, \emph{ECG5000}, and \emph{ElectricDevices}, which have a relatively large number of samples, and \emph{CinCECGTorso}, \emph{Mallat}, and \emph{Phoneme}, which contain relatively long time series.
Each data point represents the cumulative time required for fitting the parameters of \castor{} and for transforming the samples. To highlight the underlying trend, we fit a second-degree polynomial, represented as a dashed line, to the scalability data of each dataset. We observe that, as anticipated, the computational cost increases linearly with both the number of samples and the number of time steps. Furthermore, when we concurrently scale the number of samples and the number of time steps (middle), it is evident that the scalability of \castor{} follows the anticipated pattern, with the computational cost scaling in proportion to $O(nm)$.

\subsection{Ablation study}\label{sec:ablation}
In this section, we investigate the principal hyperparameters of \castor{} as follows: () the effect of competition measured by the balance between the number of groups ($g$) and the number of shapelets per group ($k$), (2) the selection of counting strategies for the minimum and maximum distances, (3) the decision to use independent or competing occurrences, (4) the implementation of first-order differences, and (5) the effects of $z$-normalization. To ensure that the conclusions regarding the general performance of \castor{} remain valid, we use a subset of 40 datasets from the UCR repository as a \emph{development set}. The development set is used to evaluate the performance of algorithmic choices, such as the number of groups, the number of shapelets, or the normalization probability.

We perform all experiments in the sensitivity analysis using 5-fold cross validation. All results are the mean over 5 runs of 5-fold cross validation.

\begin{figure}
	\begin{center}
		\includegraphics[]{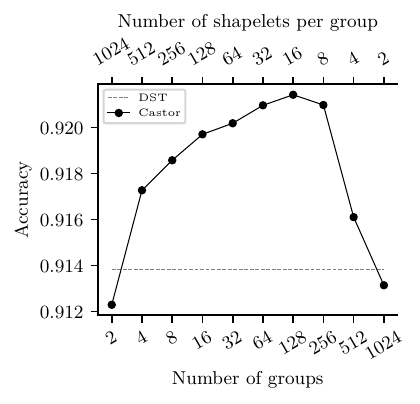}
	\end{center}
	\caption{The average accuracy, measured on the development set, for pairs of $g$ and $k$ such that $g \times k = 2048$ compared to the baseline DST, which approximately corresponds to $g=1$, $k=10000$.}\label{fig:n_groups-n_shapelets}
\end{figure}

\subsubsection{The number of groups and shapelets}\label{sec:n_groups-n_shapelets}
To investigate the effect of \emph{competition} among shapelets on the performance of \castor{}, we varied the number of groups $g$ and the number of shapelets per group $k$. We set $g \in \{2^1, 2^2, \ldots, 2^{10}\}$ and $k \in \{2^{10}, 2^9, \ldots, 2^1\}$, while maintaining the product $g \times k$ at a constant value of 2048.
In the scenario with minimal competition, where $g=1024$ and $k=2$, \castor{} is analogous to DST with $2048 \times (\log_2(m-l) + 1)$ shapelets.
Conversely, in the scenario with maximal competition, characterized by $g=2$ and $k=1024$, \castor{} is akin to a dictionary-based transformation that utilizes shapelets as counted patterns.
Figure~\ref{fig:n_groups-n_shapelets} illustrates the average performance across all development datasets for each combination of $g$ and $k$, with the performance of $g$ and $k$ corresponding to the bottom and top x-axes, respectively. The baseline performance of DST is depicted as a gray dashed line.
The results indicate that extreme competition settings such as full or no competition do not surpass the baseline performance of DST.
However, a notable enhancement in predictive performance is observed when the number of groups and the number of shapelets per group are balanced.
The optimal accuracy on the development set is achieved when the number of groups is eight times the number of shapelets.
Therefore, we recommend default values of $g=128$ and $k=16$.  

\begin{figure}
	\begin{center}
		\subfloat[Hard vs. soft minimum and maximum distances under independent or competing occurrence.]{\label{fig:hard-soft-min-max-threshold}
			\includegraphics[width=0.475\textwidth]{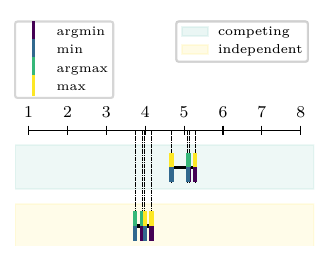}
		}
		\subfloat[Lower and upper threshold determination.]{\label{fig:lower-upper}
			\includegraphics[width=0.475\textwidth]{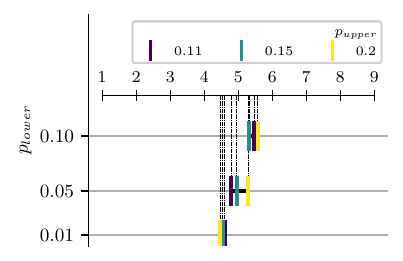}
		}
	\end{center}
	\caption{Comparison of using hard or soft minimum and maximum distance, conditioned on independent or competing occurrence (left) and the choice of the lower and upper threshold determination value (right).}\label{fig:hard-soft-lower-upper}
\end{figure}

\subsubsection{Competing features and subsequence occurrence}
In Figure~\ref{fig:hard-soft-min-max-threshold}, we investigate the impact of hard and soft constraints for minimum and maximum distances, as well as the effect of competing and independent occurrences.
Figure~\ref{fig:hard-soft-min-max-threshold} depicts the average rank associated with the combination of hard/soft minimum/maximum distance constraints as colored vertical lines, set against the backdrop of colored horizontal regions that represent independent (light yellow) and competing (light green) occurrences.

First, the highest-ranked combination with competing occurrences is inferior in rank to the lowest-ranked combination with independent occurrences, which makes the competing occurrences preferred as a recommend option for \castor{}.
Second, despite the rankings being very similar and lacking significant differences, the combination that combines a soft minimum (green) and a hard maximum (blue) constraint achieves the highest rank.
Therefore, we advocate for the default use of independent occurrences in conjunction with a soft minimum distance and a hard maximum distance.

\subsubsection{Lower and upper occurrence threshold}
In Figure~\ref{fig:lower-upper}, we examine the effects of varying the lower and upper probabilities for determining the occurrence threshold.
The figure illustrates the average rank associated with the upper probability $\rho_{\text{upper}}$ as vertical colored bars, with the lower probability specified on the y-axis.
Moreover, a discernible trend suggests that an increase in $\rho_{\text{lower}}$ leads to a deterioration in the average rank.
For the range of upper probabilities examined, there is no definitive trend indicating that expanding the range significantly influences the selection of the occurrence threshold.
Specifically, at $\rho_{\text{lower}}=0.01$, the average performance remains nearly unchanged regardless of the selected upper bound.
Consequently, we opt to use the largest range with $\rho_{\text{lower}}=0.01$ and $\rho_{\text{upper}}=0.2$ as the default parameters for \castor{}.

\subsubsection{Subsequence normalization}
\begin{figure}
	\begin{center}
		\subfloat[Subsequence length under different levels of normalization probability.]{\label{fig:normalize-prob-shapelet-size}
			\includegraphics[width=0.475\textwidth]{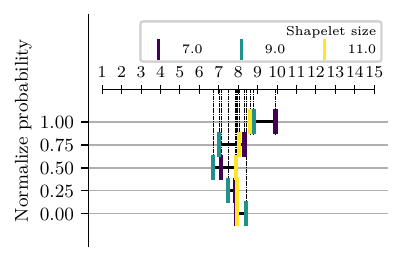}
		}
		\subfloat[First-order differences and number of groups.]{\label{fig:n-groups-diff}
			\includegraphics[width=0.475\textwidth]{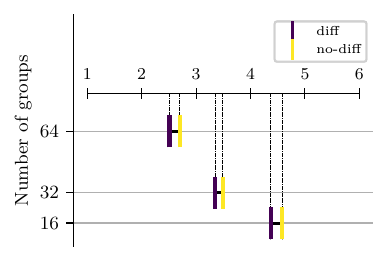}
		}
	\end{center}
	\caption{Comparison of the effect of $\rho_{\text{norm}}$ and subsequence size (left) and first-order differences and number of groups (right).}\label{fig:cmp1}
\end{figure}

In Figure~\ref{fig:normalize-prob-shapelet-size}, we investigate the impact of the proportion of groups utilizing $z$-normalized distance profiles $\rho_{\text{norm}}$ and the subsequence length $l\in \{7, 9, 11\}$ on the average performance across the development datasets.
The selection of subsequence lengths was informed by the recommendations presented in \citet{guillaume2022randp}.
The figure conveys the average rank associated with each combination of $\rho_{\text{norm}}$ and $l$, with the normalization probability depicted on the y-axis and each subsequence length represented by a distinct vertical colored line.
It is evident that, except for the two extreme cases where normalization is either always or never applied, a subsequence length of $l=9$ yields the most favorable average rank. Indeed, for these levels of normalization probability, the longest subsequence length among the three considered consistently achieves the best average rank.
Contrary to expectations, the strategy of always using $z$-normalized distance profiles underperforms relative to the approach of never using normalized profiles.
Nonetheless, the results suggest that the highest-ranked normalization probability and subsequence length is situated between the two extremes strategies, specifically at $\rho_{\text{norm}}=0.5$ and $l=9$.
Therefore, the default parameters for \castor{} are set to apply $z$-normalized distance profiles to $50\%$ of the groups with a subsequence length of $9$.

\subsubsection{First-order differences}
Figure~\ref{fig:n-groups-diff} illustrates the performance difference when employing first-order differences for $50\%$ of the groups, as opposed to relying solely on the original time series data.
The figure demonstrates the effect of applying first-order differences, represented by colored vertical lines, across various numbers of groups $g \in \{16, 32, 64\}$, with the corresponding average rank depicted on the y-axis.
The result suggests that the use of first-order differences has a positive effect on the average rank, regardless of the number of groups.
Given that first-order differences positively influence performance on the development datasets, they have been adopted for half of the groups as a component of the default parameters.

\section{Conclusion}\label{sec:conclusion}
We presented \castor{}, a state-of-the-art shapelet-based time series classifier that transforms time series data into a discriminative feature representation using shapelets with distance-based statistics. \castor{} builds upon the concept of \emph{competition}, and adapts it for use with shapelet-based transformations. The algorithm facilitates a balance between competition and independence among shapelets, yielding a transformation that achieves state-of-the-art predictive accuracy when compared against state-of-the-art algorithms and represents the best published method for shapelet-based time series classification.

In future work, sophisticated ensemble methods based on \castor{} can be explored including their application to multivariate time series analysis.
Considering that \castor{} utilizes inherently interpretable components, the explainability of classifiers that are based on the \castor{} transformation can be explored. Furthermore, we aim to assess the predictive capabilities of \castor{} when using elastic distance measures and to identify additional innovative features that can be extracted from the \castor{} parameters. Finally, we aim to apply \castor{} to other tasks such as clustering or extrinsic time series regression.

\backmatter

\bmhead{Supplementary information}

We provide the full source-code for Competing Dilated Shapelet Transform and all state-of-the-art methods that were used in the experiments to simplify reproducibility. We also provide complete results for all repetitions of the experiments at the supporting website.




\section*{Declarations}
The source code is available on the supporting website.
Isak Samsten received project funding from Vetenskapsrådet VR-2023-01588.

\bibliography{references}

\end{document}